\documentclass[conference]{IEEEtran}
\IEEEoverridecommandlockouts
\usepackage{cite}
\usepackage{amsmath,amssymb,amsfonts}
\usepackage{algorithmic}
\usepackage{graphicx}
\usepackage{textcomp}
\usepackage{xcolor}
\usepackage[colorlinks,
            linkcolor=red,
            anchorcolor=black,
            citecolor=red]{hyperref}
\def\BibTeX{{\rm B\kern-.05em{\sc i\kern-.025em b}\kern-.08em
    T\kern-.1667em\lower.7ex\hbox{E}\kern-.125emX}}
\DeclareRobustCommand*{\IEEEauthorrefmark}[1]{%
    \raisebox{0pt}[0pt][0pt]{\textsuperscript{\footnotesize\ensuremath{#1}}}}
\begin{document}
\title{HybridPoint: Point Cloud Registration Based on Hybrid Point Sampling and Matching}

\author{
\IEEEauthorblockN{
Yiheng Li\IEEEauthorrefmark{1},
Canhui Tang\IEEEauthorrefmark{1},
Runzhao Yao\IEEEauthorrefmark{1},
Aixue Ye\IEEEauthorrefmark{2},
Feng Wen\IEEEauthorrefmark{2}, and
Shaoyi Du\IEEEauthorrefmark{1,*}}
\IEEEauthorblockA{\IEEEauthorrefmark{1}National Key Laboratory of Human-Machine Hybrid Augmented Intelligence, \\ National Engineering Research Center for Visual Information and Applications, \\and Institute of Artificial Intelligence and Robotics, Xi’an Jiaotong University, Xi’an, China}
\IEEEauthorblockA{\IEEEauthorrefmark{2}Huawei Noah's Ark Lab, Beijing, China}
\IEEEauthorblockA{\IEEEauthorrefmark{*}Corresponding author, Email: dushaoyi@gmail.com}}

\maketitle

\begin{abstract}
Patch-to-point matching has become a robust way of point cloud registration. However, previous patch-matching methods employ superpoints with poor localization precision as nodes, which may lead to ambiguous patch partitions. In this paper, we propose a HybridPoint-based network to find more robust and accurate correspondences. Firstly, we propose to use salient points with prominent local features as nodes to increase patch repeatability, and introduce some uniformly distributed points to complete the point cloud, thus constituting hybrid points. Hybrid points not only have better localization precision but also give a complete picture of the whole point cloud. Furthermore, based on the characteristic of hybrid points, we propose a dual-classes patch matching module, which leverages the matching results of salient points and filters the matching noise of non-salient points. Experiments show that our model achieves state-of-the-art performance on 3DMatch, 3DLoMatch, and KITTI odometry, especially with 93.0\% Registration Recall on the 3DMatch dataset. Our code and models are available at \href{https://github.com/liyih/HybridPoint}{https://github.com/liyih/HybridPoint}.
\end{abstract}


\section{Introduction}
\label{sec:intro}
Point cloud registration is an essential task in computer vision. Its purpose is to find a rigid transformation matrix to align two partially overlapping point clouds. It has a wide range of applications for SLAM, 3D reconstruction, and other fields. In recent years, with the emergence of advanced 3d sensors and the rise of deep learning, the accuracy of the point cloud registration algorithm has been rapidly improved. Despite this, there are still many difficulties in point cloud registration. For example, it is still challenging to register low overlapping point clouds and point clouds with many outlier points.\par
\begin{figure}[t]
    \centering
    \includegraphics[height=6.5cm, width=8.3cm]{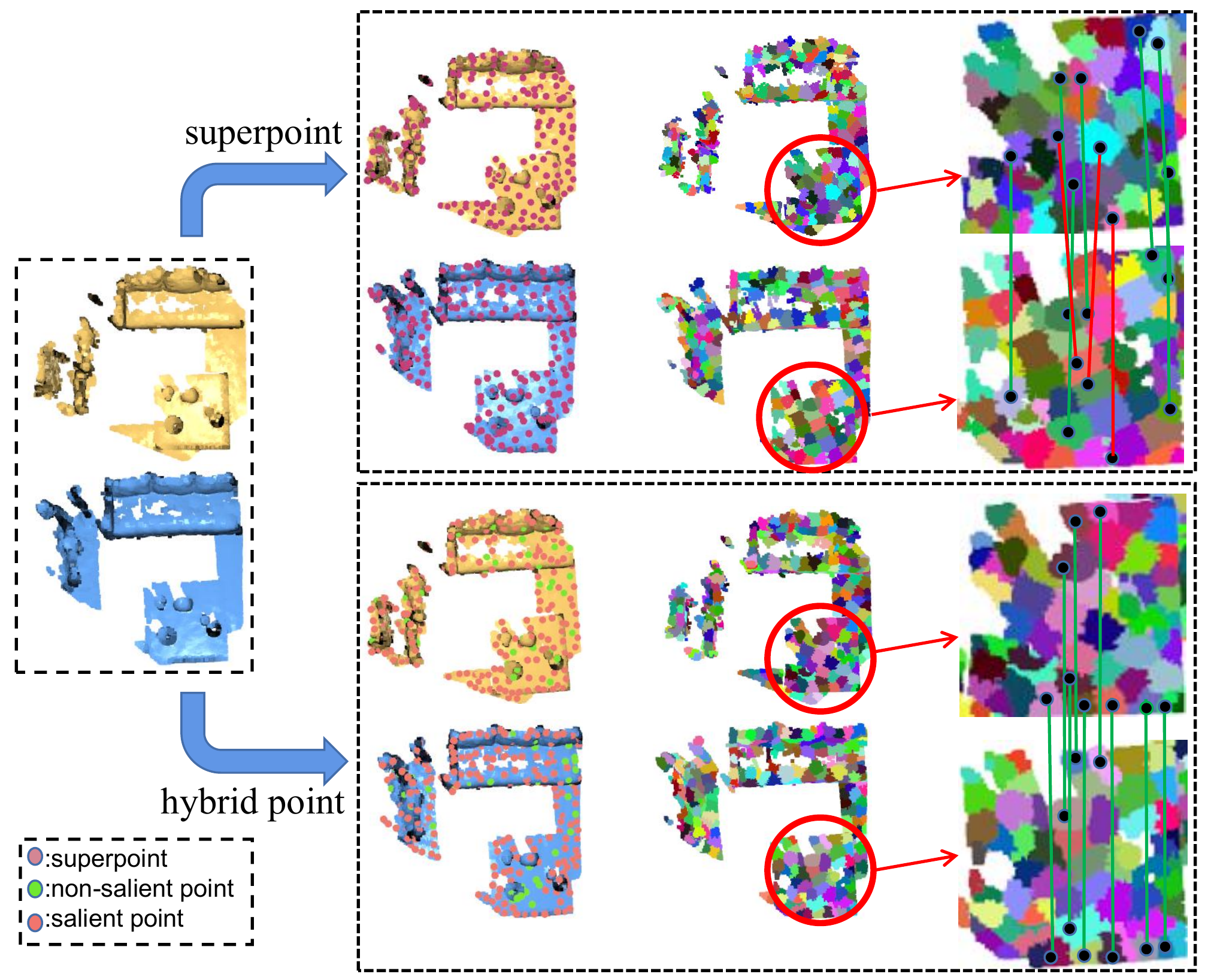}
    \caption{Given the result of point cloud pairs(left) sampled by grid sampling(right top) and hybrid point sampling(right bottom), we show the result of point-to-node grouping based on superpoints and hybrid points, respectively. We randomly select a local region and link the rightly matched patches with green lines and the wrong ones with red lines. Notice that corresponding patches have better partition consistency based on hybrid points.}
    \label{fig:1}
\end{figure}
The key to the success of feature-based point cloud registration lies in the establishment of robust correspondences with distinguishable descriptors. Recently, there have been some methods~\cite{qin2022geometric, yu2021cofinet} that extract the correspondences in a patch-to-point way. These methods first divide the original point cloud pairs into small patches and match them. Afterward, the dense points in the corresponding patches are matched to obtain the dense point correspondences. Finally, the transformation matrix is estimated based on the point correspondences. 
It is worth mentioning that these patch-to-point and keypoint-free methods perform outstandingly in low overlapping scenarios. However, their coarse patch partition produces a new problem that the matched patches are still of low overlap. In particular, the partition of patches mainly takes the grid-sampled superpoints as nodes and divides patches through the point-to-node grouping strategy~\cite{qi2017pointnet++, li2018so}. Since the grid-sampled superpoints are sparse and loose, i.e. they are not repeatable, their local neighborhood (patches) are inconsistent between the point cloud pair, which causes a bad impact on the subsequent point matching.
At the same time, due to the low saliency of superpoints, it is also problematic to use the features of superpoints to represent the features of patches for patch matching.\par

To this end, we propose to extract points with well-designed distribution and high repeatability as nodes. Depending on whether its feature is locally prominent, the points in a point cloud can be cast into two categories: salient and non-salient. Salient points, such as corner and edge points, can be easily located from different perspectives, which means they have relatively good localization precision.
More importantly, locally salient features contribute to robust patch matching.
Furthermore, to correctly and effectively conduct subsequent patch division, the extracted points should not be too dense, so a Non-Maximum Suppression module is used to give the proper density of salient points. For non-salient points, such as plane points, although they have non-significant features and poor localization precision, should not be discarded for the reason that these non-salient points are essential in registering complex cases where the overlapping parts of the point cloud pairs are mainly planes. Thus some uniformly distributed points are introduced for patch matching and partition in such areas. Finally, the union of salient non-salient points forms the hybrid points, replacing the superpoints used in the previous methods~\cite{qin2022geometric,yu2021cofinet,bai2020d3feat, huang2021predator}.

On the other side, patch matching is inevitable to fuzzy matching.
For our patch-matching stage, since hybrid points comprise salient and non-salient points representing different geometric structures in geometric space, we propose to match the two different kinds of points separately and then merge the matching results. Especially, non-salient points have non-significant features and poor localization precision, thus exiting many incorrect correspondences in their matching results.  Therefore, a spectral technique\cite{leordeanu2005spectral} is adopted to conduct additional filtering in non-salient branch. Evaluated on both 3DMatch\cite{zeng20173dmatch}, 3DLoMatch\cite{huang2021predator} and KITTI odometry~\cite{geiger2012we}, our HybridPoint shows superiority in the registration robustness and acuracy, 
especially for 3DMatch benchmark~\cite{zeng20173dmatch} with 93.0$\%$ Registration Recall.
Our major contributions are as follows:\par
\begin{itemize}
\item We propose to extract salient points at the patch phase to solve the problem of inconsistent patch partition that previous works overlook. Consequently, it makes the matched patches have
higher overlaps and benefits for the subsequent point matching.
\item We propose to select non-salient points to give a full picture of the whole point cloud, which is necessary for those hard cases where the overlapping region has only a small number of salient points. The union of salient and non-salient points is hybrid points.
\item We propose a dual-classes patch matching module, which leverages the matching results in salient regions and filters the matching noise in non-salient regions, contributing to more robust registration results. 
\end{itemize}

\section{Method}

\subsection{Overview}
\begin{figure*}[htp]
    \centering
    \includegraphics[height=6.7cm, width=18cm]{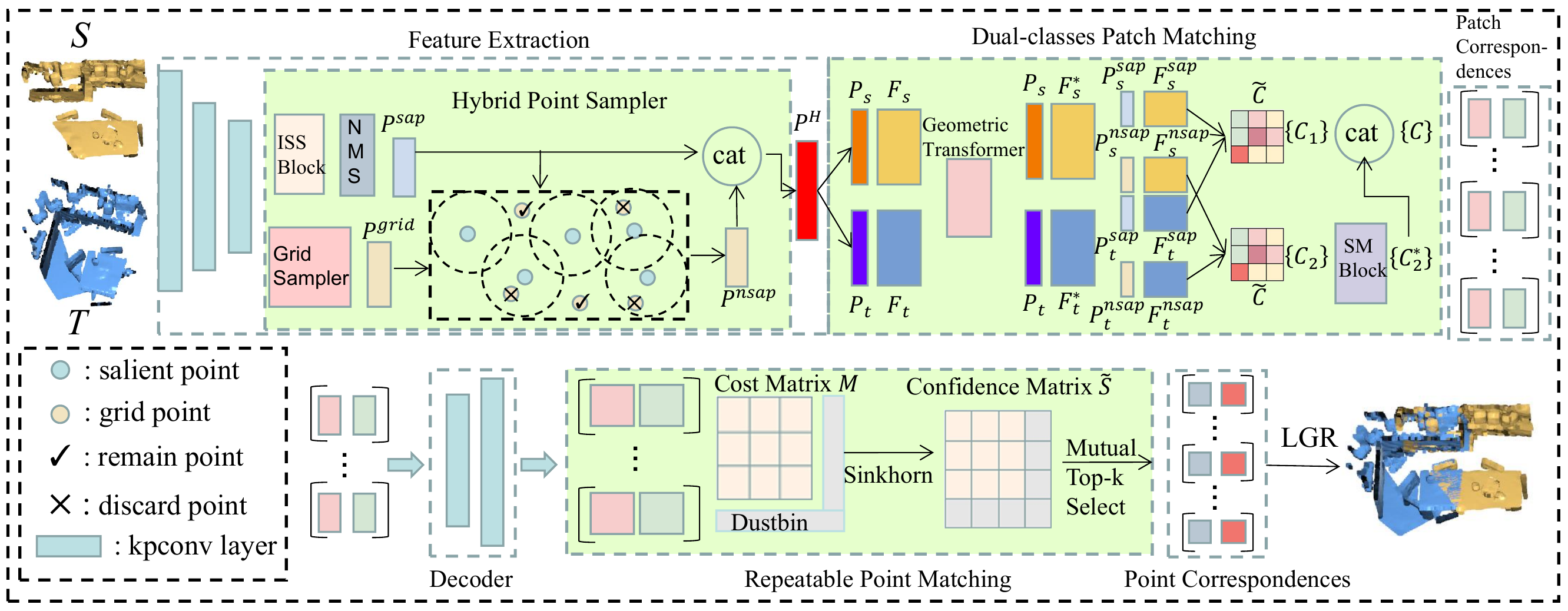}
    \caption{Given two partially overlapping point clouds $S$ and $T$, we first adopt KPConv-FPN~\cite{thomas2019kpconv} to learn point-level features. At the last layer of the encoder, we propose a Hybrid Point Sampler to extract hybrid points (including salient points ${P}^{sap}$ and non-salient points ${P}^{nsap}$) instead of original superpoint\cite{qin2022geometric,huang2021predator,yu2021cofinet}. Then, a Dual-classes Patch Matching module is proposed to extract patch correspondences $\{C\}$ through the way, remaining salient correspondences $\{C_1\}$ and filtering non-salient correspondences $\{C_2\}$ with the spectral matching algorithm. The Repeatable Point Matching Module extracts point correspondences from matched patch pairs based on the confidence matrix $\Tilde{S}$. Finally, the local-to-global registration(LGR) method estimates the transformation matrix.} 
    \label{fig2}
\end{figure*}

Our model adopts the patch-to-point method, which finds dense point correspondences in a coarse-to-fine manner. As shown in Fig.\ref{fig2}, firstly, the Feature Extract module downsamples the input point clouds in multiple resolution levels and learns point-level features. During learning features, the points at the end of the encoder are replaced by hybrid points sampled by hybrid points sampler based on the last but one resolution level. Next, the dual-classes patch matching module, which leverages the matching results in salient regions and filters the matching noise in non-salient regions, extracts patch correspondences. Then, dense point correspondences are generated based on matched patches through repeatable point matching. Finally, the local-to-global registration method estimates the transformation matrix.

\subsection{Hybrid point}
Previous patch-to-point matching methods~\cite{qin2022geometric, yu2021cofinet} use superpoints with poor localization precision as nodes, leading to low overlapping matched patches. Points with significant local features could be the candidates for nodes as they can be easily localized in different perspectives. However, such points distribute sparsely on the plane, which leads to registration failure in some challenging cases. Therefore, in the area that salient points cannot describe, some uniformly distributed points are introduced, combing them with salient points to form hybrid points.\par
\textbf{Salient point: }
Our model adopts the grid sampler to sample multi-level resolution points from the original point cloud $P$. The resolution of these sampled point clouds decreases consecutively. Salient points are extracted from the last but one resolution level $P_2 =\{\bf{p}_i\in \mathbb{R}^3|i=1,...,N\}$ by combining Intrinsic Shape Signatures~\cite{zhong2009intrinsic} with Non-Maximum Suppression. Building a local coordinate system for each point $p_i$ in $P_2$, then using radius ${r}$ search ${K}$ nearest neighbor $N =\{\bf{n}_j\in \mathbb{R}^3|j=1,...,K\}$ for  $p_i$, lastly calculating the weight of every point in the set ${N}$ according to the distance as follows:

\begin{equation}\label{eq2}
{w}_{j} = 1 / | {p}_{i} - {p}_{j}| , | {p}_{i} - {p}_{j}|<{r}.
\end{equation}

Getting corresponding weight set $W =\{\bf{w}_j\in \mathbb{R}^1|j=1,...,K\}$ of set ${N}$. The covariance matrix $m$ is calculated through weight set ${W}$ and $K$ nearest neighbor set ${N}$ based on the formulation: 

\begin{equation}\label{eq2}
\!cov({p}_{i})=\sum_{|{p}_{i} - {p}_{j}|<{r}} {w}_{j}({p}_{i} - {p}_{j}){({p}_{i} - {p}_{j})}^{T} / \sum_{|{p}_{i} - {p}_{j}|<{r}}{w}_{j}\!.
\end{equation}

Executing the above operations in every point of set ${P}_{2}$ and obtaining the covariance matrix set $M =\{\bf{m}_i\in \mathbb{R}^{3\times3}|i=1,...,N\}$, the covariance matrix in ${M}$ corresponds one to one with the points in ${P}_{2}$. The eigenvalues $\{\bf{\lambda}_1\ , \bf{\lambda}_2\ , \bf{\lambda}_3\ \}$ of each covariance matrix are calculated and are ranked from largest to smallest. The threshold values $\gamma_1$ and $\gamma_2$ are set. If the following formula is satisfied, the point is considered as the point with a prominent feature, and the smallest eigenvalue among these three eigenvalues is recorded as ${v}$.  
\begin{equation}\label{eq2}
{\lambda}_{2}/{\lambda}_{1} \leq \gamma_1 ,{\lambda}_{3}/{\lambda}_{2} \leq \gamma_2.
\end{equation}

It is important to note that the above points can not be regarded as salient points directly. The reason is that the following steps will utilize salient points to divide the patch. Each point will represent an area, so the salient points should not be too dense. The non-maximum suppression is conducted on salient points set $S =\{\bf{s}_i\in \mathbb{R}^{3}|i=1,...,L\}$, based on its corresponding set of minimum eigenvalue $V =\{\bf{v}_i\in \mathbb{R}^{1}|i=1,...,L\}$. After the suppressed points are deleted from set S, the final retained points are salient points $P^{sap}$.\par

\begin{figure}[t]
    \centering
    \includegraphics[height=3cm, width=7.5cm]{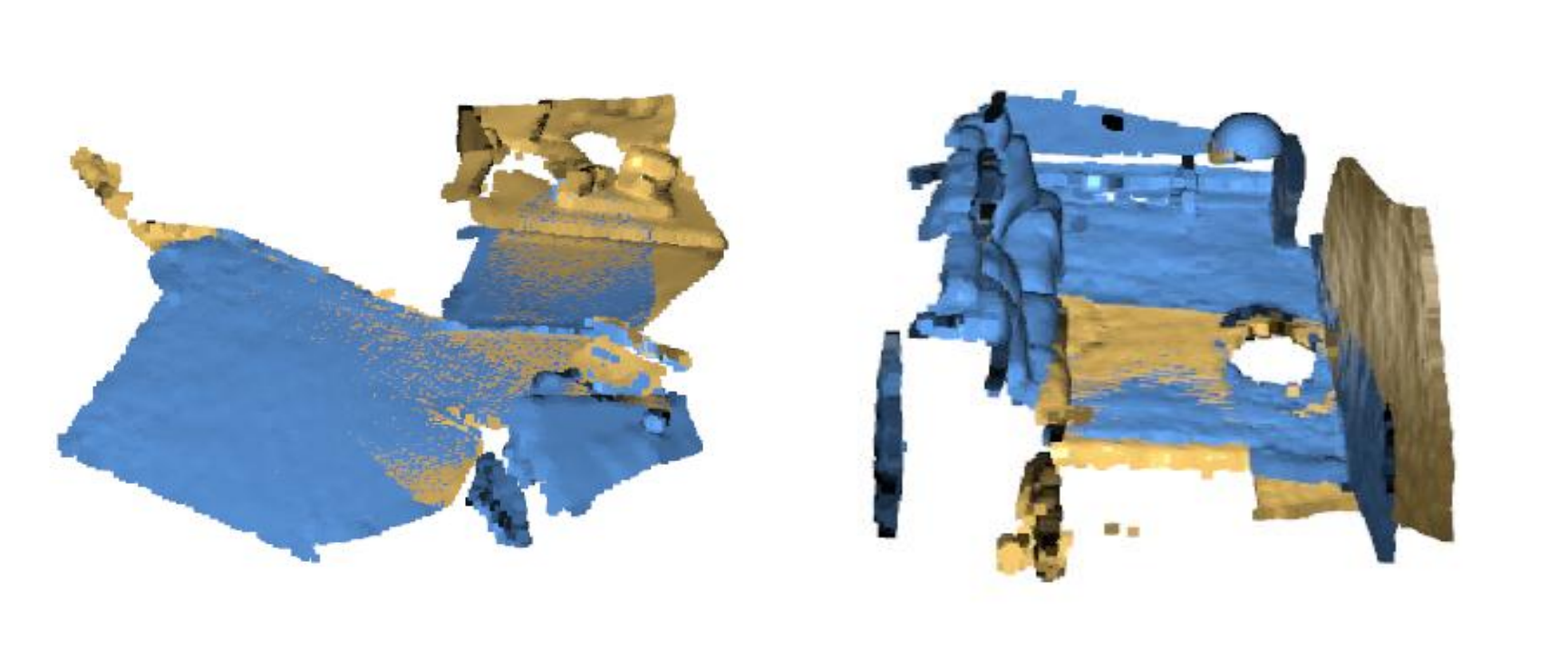}
    \caption{Two hard cases where the overlapping region of the point cloud pairs are mainly plain. In these cases, non-salient points are necessary to support enough correspondence to achieve accurate registration.}
    \label{fig11}
\end{figure}

\textbf{Non-salient point: }
To fully describe the whole point cloud, it also needs to select some uniformly distributed points in the regions that the $P^{sap}$ cannot describe. For those hard cases where the overlapping region has only a small number of salient points (see Fig. \ref{fig11}), non-salient point points are necessary for supporting enough correspondences that are essential for accurate point cloud registration. The last level downsampled point cloud $P_3$ can be approximately regarded as uniformly distributed and has a similar density to $P^{sap}$. Therefore, the points in ${P}_{3}$ that are far away from the salient points can be selected and regarded as located in non-salient regions. These points are recorded as non-salient points $P^{nsap}$.

Taking a point ${p}_{i}$ from set ${P}_{3}$. Calculating the minimum distance ${d}$ between ${p}_{i}$ and all the points in the salient points set ${P}^{sap} =\{\bf{p}_l^{sap}\in \mathbb{R}^{3}|l=1,...,L\}$. ${d}$ can be calculated by the following formula:

\begin{equation}\label{eq2}
d=\min ||{p}_{i} - {p}^{sap}_l|| , l=1,...,L.
\end{equation}

Take ${\sigma}$ as the distance threshold. If ${d}$ $>$ ${\sigma}$, it can be considered that the region in which ${p}_{i}$ located cannot be described by the salient points; otherwise, it can be considered that the salient points can describe the region. All the points locate far away enough from salient points form the non-salient points set ${P}^{nsap} =\{\bf{p}_t^{nsap}\in \mathbb{R}^{3}|t=1,...,T\}$.The union of ${P}^{sap}$ and ${P}^{nsap}$ is the hybrid points set ${P}^{H}$.

\subsection{Dual-classes patch matching}
Due to the significant difference in the spatial distribution of the two types of points, the two types of points have different matching tendencies, i.e. salient points tend to match with salient points, and non-salient points tend to match with non-salient points, we propose to match the two type of points separately. What's more,
filtering out mismatched patch correspondences helps to get more robust point correspondences at the point matching~\cite{qin2022geometric} stage. However, due to the sparse and loose nature of patch matching, many proper correspondences are often suppressed during filtering. We propose to use the idea of divide-and-conquer when filtering and give different filter strategies for different types of patch-feature matching results. As salient points are easier to match correctly (76.5$\%$ IR) while most of the non-salient points cannot be matched (24.5$\%$ IR). Therefore, we extract salient points to match them first and find top 10$\%$ reliable correspondences matched by non-salient points as a supplement. 

\textbf{Transformer architecture: }Transformer architecture~\cite{vaswani2017attention} can effectively improve the size of the receptive field and obtain more distinguishable descriptors. Therefore, the geometric transformer~\cite{qin2022geometric} are adopted for the interaction process. Geometric transformer proposes geometric structure embedding that can obtain geometric structure features with rotation invariance. 
The features associated with the end of the encoder, denoted by ${F}_{S} \in \mathbb{R}^{|S|\times d}$ and ${F}_{T} \in \mathbb{R}^{|T|\times d}$, are the input of the geometric transformer. The output of the geometric transformer is represented by 
${F}_{S}^{*} \in \mathbb{R}^{|S|\times d}$ and 
${F}_{T}^{*} \in \mathbb{R}^{|T|\times d}$,
where ${|S|}$ and ${|T|}$ refer to the number of hybrid points in the source point cloud and target point cloud, respectively.\par

\textbf{Hybrid point matching: }
Since the distribution of salient and non-salient points in geometric space is significantly different, we propose to match salient and non-salient points separately and denote the patch matching results by ${\{C_1\}}$ and ${\{C_2\}}$, respectively.
Each matching process are mainly as follows:

Given a subset from ${F}_{S}^{*}$ and ${F}_{T}^{*}$, respectively, denoting them by ${F}_{s}^{*}$ and ${F}_{t}^{*}$. Firstly, the correlation matrix $\Tilde{C} \in \mathbb{R}^{|s|\times |t|}$ are calculated based on normalized features $\Tilde{F}_{s}^{*}$ and $\Tilde{F}_{t}^{*}$. Then, we also utilize a dual-normalization operation on $\Tilde{C}$ to suppress ambiguous matches. Finally, the largest ${K}$ entries in $\Tilde{C}$ are selected as the patch correspondences.

To retrieve as many as inliers from initial patch correspondences, we employ the spectral matching algorithm~\cite{leordeanu2005spectral} to filter the correspondences in ${\{C_2\}}$, the reason why the SM is only used in ${\{C_2\}}$ is shown in section \ref{sec:explation}. Firstly, a compatibility matrix is computed for ${\{C_2\}}$ based on the 3D spatial consistency, in which each element is the pairwise affinity score of two correspondences. Secondly, the principle eigenvector of the compatibility matrix is calculated through eigen analysis. Thirdly, locate the maximum element $e$ of the principle eigenvector, and remove the component that conflicts with the corresponding item of $e$ from $\{C_2\}$. Finally, repeat the previous step until $e$ = 0 or $|\{C_2\}|$ equals the least number of the main cluster. The main cluster is denoted by ${\{C_2^{*}\}}$, and the union of $|\{C_1\}|$ and ${\{C_2^{*}\}}$ is the final patch
correspondences denoted as ${\{C\}}$.

\subsection{Repeatable point matching}

The Repeatable Point Matching Module is proposed to extract the dense point correspondences based on our patch correspondences. LGR~\cite{qin2022geometric} obtains the candidate matrix through the point correspondence generated by each matching patch pair and then selects the global optimal transformation matrix. Our highly overlapping patch correspondences can generate more robust and repeatable point correspondences in local regions, contributing to better candidate matrices.

For each patch correspondences ${C}_i=({P},{Q})$, firstly, compute the salient cost matrix $M^{sap} \in \mathbb{R}^{m\times n}$ and non-salient cost matrix $M^{nsap} \in \mathbb{R}^{m\times n}$ as follow: 
\begin{align}
    M^{sap}_i&=(\Tilde{F}^{sap}_P)(\Tilde{F}^{sap}_Q)^T/\sqrt{d},\\
    M_i^{nsap}&=(\Tilde{F}^{nsap}_P)(\Tilde{F}^{nsap}_Q)^T/\sqrt{d}
\end{align}
\label{eq2}
where $m$ and $n$ denotes the number of points in $P$ and $Q$, respectively. If the patch correspondence is salient, the first equation is adopted; otherwise, the second one is adopted. Then, we utilize a learnable dustbin parameter $\alpha$ and Sinkhorn algorithm~\cite{sinkhorn1967concerning} to compute the soft assignment matrix $\Tilde{S}_i$ as the confidence matrix. Finally, extracting point correspondences through mutual
top-K selection, where the point pair is selected if its corresponding entries of matrix $\Tilde{S}_i$ is the k largest in both row and column.

\section{Experiment}
\subsection{Implentation details}

Overlap-aware circle loss~\cite{sun2020circle, qin2022geometric} and Point matching loss~\cite{qin2022geometric} are employed as our loss functions for the supervision of patch matching and point matching, respectively. We utilize the Adam optimizer~\cite{loshchilov2017decoupled} with an initial learning rate of 0.0001 and weight decay of 0.000001. Step learning is used with a learning rate decay of 0.95. 40-epoch and 160-epoch training is conducted on the 3Dmatch dataset and KITTI odometry, respectively. In the hybrid points sampler, the threshold values $\gamma_1$ and $\gamma_2$ are both 0.6. In dual-classes patch matching, the inlier threshold is set to 0.1 when using Spectral matching, and the top 10\% points of confidence are retained. 
In addition, the distance threshold $\sigma$ and radius $r$ both depend on the resolution of the input data, and we set $\sigma$ to 0.15m and $r$ to 0.15m for 3DMatch and 3DLoMatch. For KITTI, we set $\sigma$ to 3.0m and $r$ to 3.0m.
Our model is trained on two NVIDIA RTX 3090 GPUs.

\subsection{3DMatch \& 3DLoMatch}
\label{sec:explation}
\textbf{Datasets : }3DMatch~\cite{zeng20173dmatch} is an indoor point cloud registration dataset obtained by RGBD image reconstruction. It consists of 62 scenes in total. According to the partitioning method of protocols~\cite{zeng20173dmatch}, 58 scenes are selected as the training set and 8 scenes as the testing set. 3DLomatch~\cite{huang2021predator} is a more difficult dataset derives from 3DMatch. In the original 3DMatch, only point cloud pairs with an overlapping rate greater than 30\% are used for testing, while the testing set of 3DLomatch included point cloud pairs with an overlapping rate between 10\% and 30\%.

\textbf{Metrics: }
The following metrics are utilized on 3Dmatch and 3DLomatch to evaluate the performance of our model. (1) Inlier Ratio (IR): Under the transformation of ground truth, the fraction of point correspondences whose distance is smaller than a certain threshold (i.e.0.1 m). (2) Feature Matching Recall (FMR): the fraction of point cloud pairs whose Inlier Ratio is above a certain threshold(i.e.5\%). (3) Registration Recall (RR): the fraction of the estimated transformation matrix whose error is smaller than a certain threshold(RMSE$<$0.2). (4) Relative Rotation Error (RRE) and Relative Translation Error (RTE): measuring the error of the successfully estimated transformation matrix compared to the ground truth.

\textbf{Registration results: }Our model is compared with FGFH~\cite{choy2019fully}, D3feat~\cite{bai2020d3feat}, Predator~\cite{huang2021predator}, Cofinet~\cite{yu2021cofinet}, and Geotransformer~\cite{qin2022geometric}. 
For Geotransformer~\cite{qin2022geometric}, we compare the effect of using LGR~\cite{qin2022geometric} as an estimator of the transformation matrix instead of RANSAC, as LGR is a faster and more robust method than RANSAC.
For all the method, we use 1000 correspondences for evaluation. As shown in Table \ref{tab1}, in the 3DMatch dataset, our model achieves the best performance in all five metrics, and the 3DLomatch model achieves the best performance in all four metrics except FMR and the second-best performance in FMR. RR, IR, and FMR indicate that our model can extract more reliable correspondences and has a higher registration success rate compared with previous models, while RRE and RTE indicate that our model has a higher precision under the condition of successful registration.\par
\begin{table}[t]
\begin{center}
\setlength\tabcolsep{2pt}
\caption{Evaluation results of five metrics on 3DMatch and 3DLoMatch.Best performance is highlighted in bold while the second best is marked with an underline.} 
\label{tab1}
\begin{tabular}{c|ccccc}
  \hline
   & \multicolumn{5}{c}{3DMatch} \\
  Methods &RR(\%)&IR(\%)&FMR(\%)&RRE(\textdegree)&RTE(m)\\
  \hline
  FCGF~\cite{choy2019fully} & 83.3 & 48.7 & 97.0&1.949&0.066\\
  D3Feat~\cite{bai2020d3feat} & 83.4 & 40.4 & 94.5 &2.161&0.067\\
   Predator~\cite{huang2021predator} & 90.6 & 57.1 & 96.5&2.029&0.064\\
  CoFiNet~\cite{yu2021cofinet} & 88.4 & 51.9 & 98.1 &2.011&2.011\\
  GeoTransfomer+LGR~\cite{qin2022geometric} &\underline{91.5}&\underline{70.3} & \underline{97.7} &\underline{1.625}&\underline{0.053}\\
  Ours&\bf{93.0}&\bf{76.3}&\bf{98.5}&\bf{1.517}&\bf{0.050}\\
  \hline
  & \multicolumn{5}{c}{3DLoMatch} \\
    Methods &RR(\%)&IR(\%)&FMR(\%)&RRE(\textdegree)&RTE(m)\\
  \hline
  FCGF~\cite{choy2019fully} & 38.2 & 17.2 & 74.2&3.147&0.100\\
  D3Feat~\cite{bai2020d3feat} & 46.9 & 14.0 & 67.0 &3.361&0.103\\
   Predator~\cite{huang2021predator} & 61.2 & 28.3 & 76.3&3.048&0.093\\
  CoFiNet~\cite{yu2021cofinet} & 64.2 & 26.7 & 83.3 &3.280&0.094\\
  GeoTransfomer+LGR~\cite{qin2022geometric} &\underline{74.0}&\underline{43.3} & \bf{88.1} &\underline{2.547}&\underline{0.074}\\
  Ours&\bf{75.0}&\bf{46.3}&\underline{87.2}&\bf{2.467}&\bf{0.070}\\
  \hline
\end{tabular}
\end{center}
\end{table}
Our method is significantly better than previous methods~\cite{choy2019fully, bai2020d3feat, huang2021predator, yu2021cofinet, qin2022geometric} for several reasons: Firstly, the hybrid points sampler extracts hybrid points as nodes, which can highlight the locally prominent region and give a complete picture of the whole point cloud, contributing to more precise patch partition and feature matching. Secondly, the dual-classes patch matching module leverages the matching results in salient regions and filters the matching noise in non-salient regions, improving the inlier rate of feature matching. These contribute to the more robust correspondences used for estimating the transformation matrix.

\begin{table}[h]
\begin{center}
\setlength\tabcolsep{5pt}
\caption{Runtime per fragment pair averaged over 1623 test pairs of 3DMatch. The runtime is divided into the data processing time, the model inference time, the pose estimation time of LGR (pose1), and the pose estimation time of RANSAC-1k (pose2).}
\label{tab333}
\begin{tabular}{c|cccc}
  \hline
  &data(s)&model(s)&pose1(s)&pose2(s)\\
  \hline
  GeoTransfomer
&\bf{0.131}&\bf{0.098}&\bf{0.026}&\bf{0.042}\\  
Ours&0.136&0.112&\bf{0.026}&\bf{0.042}\\
  \hline
\end{tabular}
\end{center}
\end{table}

The time comparison is shown in Table \ref{tab333}.
The tests are conducted on a single NVIDIA RTX 3090 GPU.
Compared with GeoTransformer, our method obtains performance improvement, and only adds a negligible amount of computation. First, the proposed hybrid points sampler has no learning parameters thus it can be computed in the dataloader, which enjoys the acceleration of data parallelism.
Second, the proposed dual-classes matching module takes the concatenation of the salient and non-salient patch correspondences to compute together so that it does not add too much computation.

\begin{table}[t]
\begin{center}
\setlength\tabcolsep{2pt}
\caption{Ablation study of point choice and our network architecture.}
\label{tab2}
\begin{tabular}{cccc|ccccc}
  \hline
  &&&&  \multicolumn{5}{c}{3DMatch} \\
  Sup & NSap & Sap & DM&RR(\%)&IR(\%)&FMR(\%)&RRE(\textdegree)&RTE(m)\\
  \hline
   \checkmark& &  & &91.5&70.3 & 97.7 &1.625&0.053\\
   &\checkmark &&  & 85.2 &57.4&95.7&1.714&0.057\\
   & &\checkmark&  & \underline{92.6} &\underline{76.5}&\underline{98.4}&1.528&\underline{0.051}\\
    &\checkmark &\checkmark  & & 92.0&\bf{80.9}&98.3&\underline{1.518}&0.054\\
   &\checkmark & \checkmark  & \checkmark  & \bf{93.0} &76.3&\bf{98.5}&\bf{1.517}&\bf{0.050}\\
   \hline
     &&&&  \multicolumn{5}{c}{3DLoMatch} \\
  Sup & NSap & Sap & DM&RR(\%)&IR(\%)&FMR(\%)&RRE(\textdegree)&RTE(m)\\
  \hline
   \checkmark& &  & &\underline{74.0}&43.3 & \bf{88.1} &2.547&0.074\\
   &\checkmark &&  & 56.0 &24.5&72.5&2.681&0.081\\
   &&\checkmark&  & 74.2&\underline{46.4}&87.2&\underline{2.486}&\underline{0.071}\\
    &\checkmark &\checkmark  & & \ 73.1&\bf{51.5}& 87.2&2.522&0.072\\
   &\checkmark & \checkmark  & \checkmark  & \bf{75.0} &46.3& \underline{87.2} &\bf{2.467}&\bf{0.070}\\
    \hline
\end{tabular}
\end{center}
\end{table}

\begin{table}[t]
\begin{center}
\setlength\tabcolsep{2.5pt}
\caption{Ablation study of using the spectral matching.}\label{tabb3} 
\begin{tabular}{cc|ccccc}
  \hline
  \multicolumn{2}{c|}{\underline{Dual Matching}}&  \multicolumn{5}{c}{3DMatch} \\
  Sap & NSap &RR(\%)&IR(\%)&FMR(\%)&RRE(\textdegree)&RTE(m)\\
  \hline
   &  &\underline{92.4}&75.6 & \underline{98.4} &\underline{1.532}&\underline{0.051}\\
   +SM & & 91.7 &70.9&98.1&1.673&0.052\\
   +SM&+SM & 91.2 &\bf{80.7}&97.7&1.692&0.052\\
     &+SM & \bf{93.0} &\underline{76.3}&\bf{98.5}&\bf{1.517}&\bf{0.050} \\
   \hline
  \multicolumn{2}{c|}{\underline{Dual Matching}}&  \multicolumn{5}{c}{3DLoMatch} \\
  Sap & NSap &RR(\%)&IR(\%)&FMR(\%)&RRE(\textdegree)&RTE(m)\\
  \hline
   &  &\underline{74.2}&45.8 &\bf{87.4} &\underline{2.503}&\underline{0.070}\\
   +SM & & 71.1 & 41.1 &85.2&2.611&0.076\\
   +SM&+SM  &69.3&\bf{52.6}&84.0&2.716&0.077 \\
 &+SM & \bf{75.0} &\underline{46.3}&\underline{87.2}&\bf{2.467}&\bf{0.070}\\
   \hline

\end{tabular}
\end{center}
\end{table}

\begin{figure}[h]
	\centering
	\begin{minipage}{0.48\linewidth}
		\vspace{4pt}
\centerline{\includegraphics[width=\textwidth]{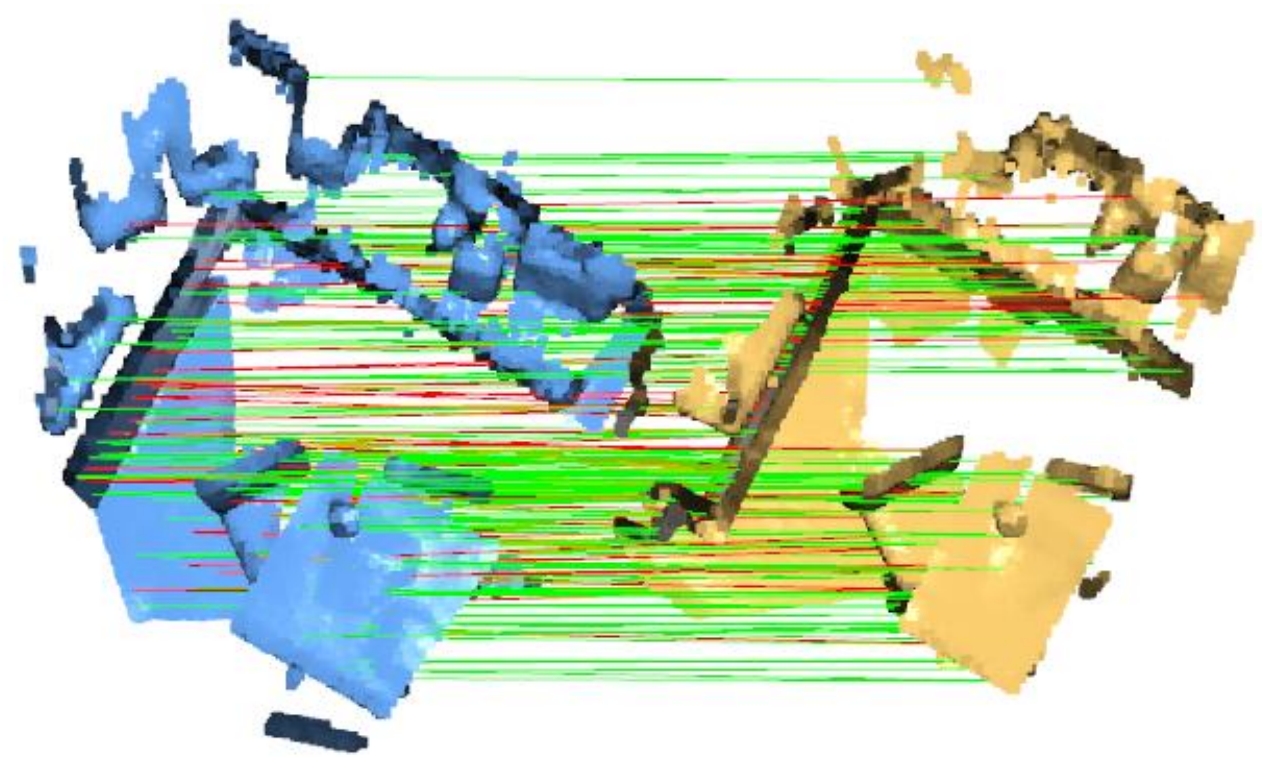}}
		\centerline{(a) superpoint}
	\end{minipage}
	\begin{minipage}{0.48\linewidth}
		\vspace{4pt}
\centerline{\includegraphics[width=\textwidth]{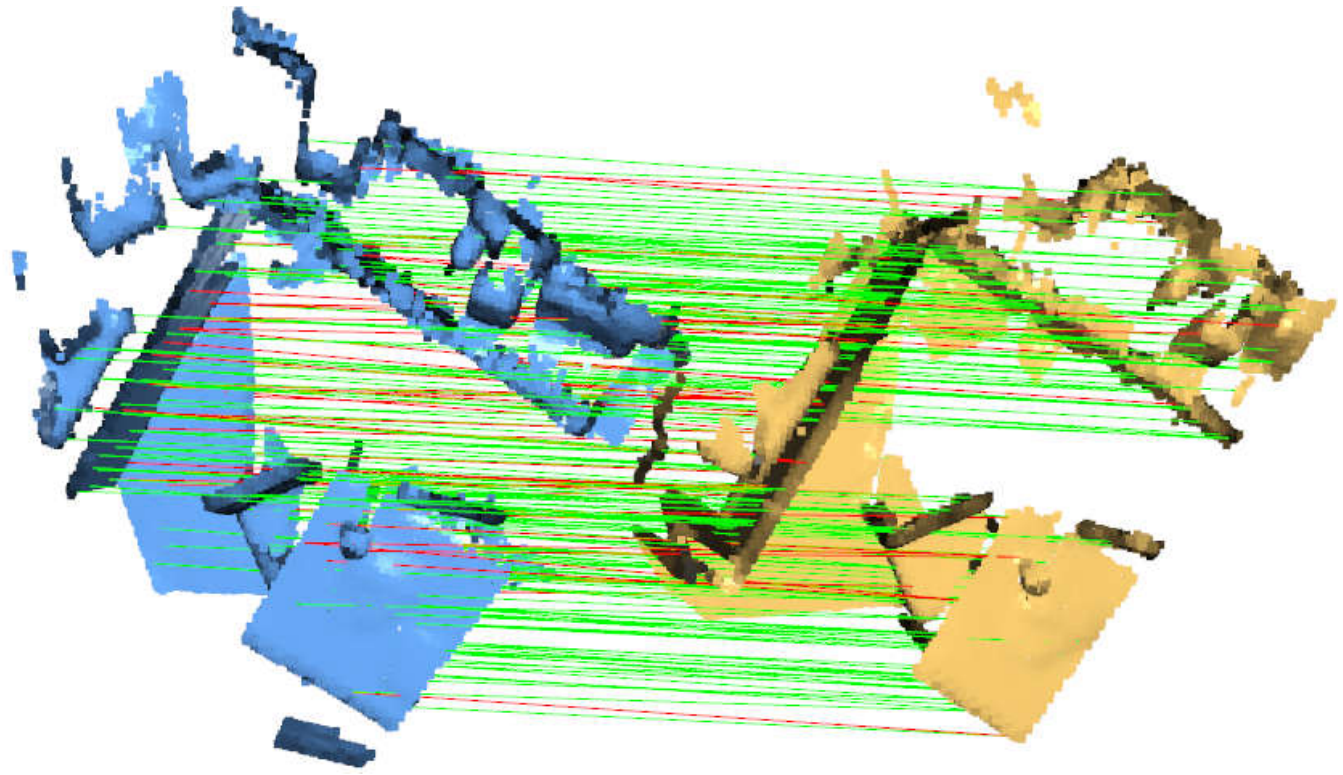}}
		\centerline{(b) salient point}
	\end{minipage}
 
	\begin{minipage}{0.48\linewidth}
		\vspace{4pt}
\centerline{\includegraphics[width=\textwidth]{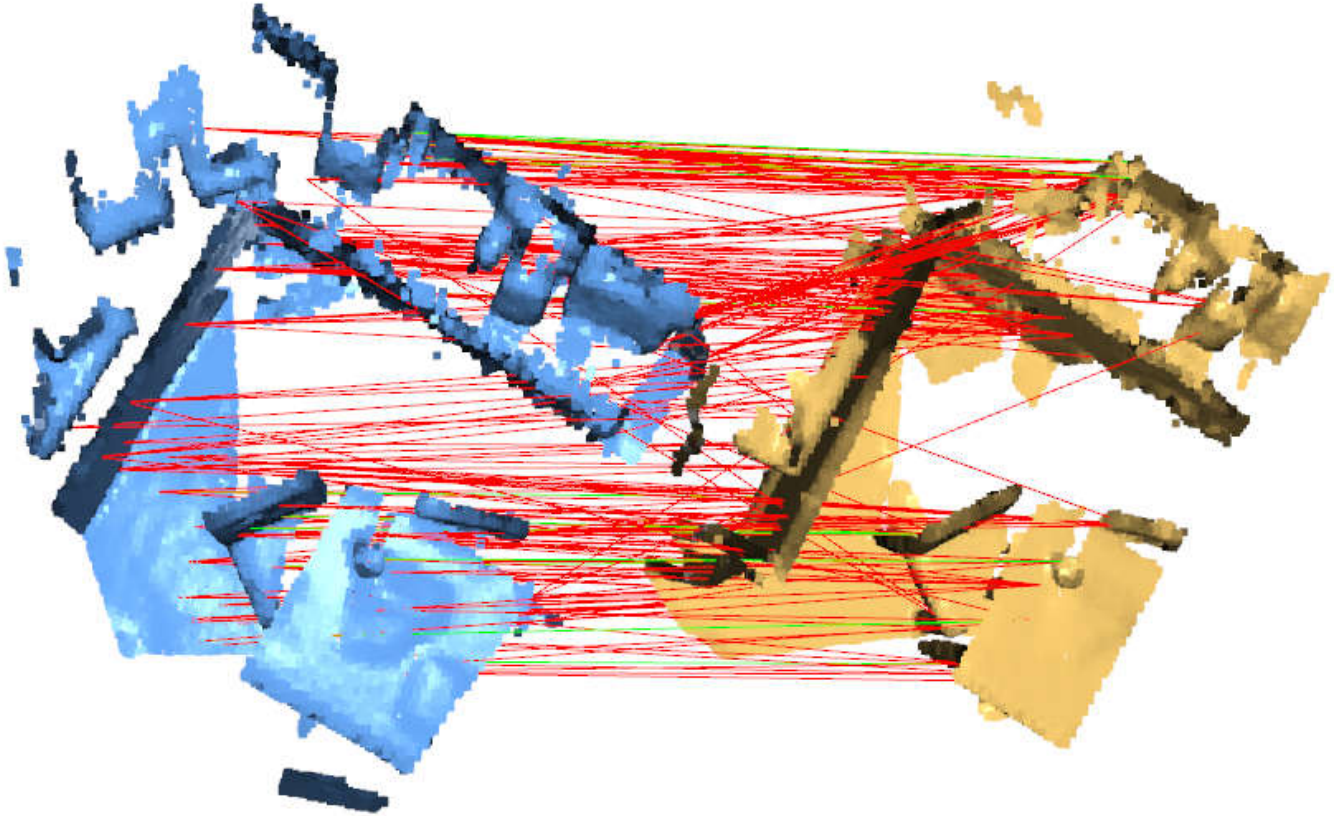}}
		\centerline{(c) non-salient point}
	\end{minipage}
        \begin{minipage}{0.48\linewidth}
		\vspace{4pt}
\centerline{\includegraphics[width=\textwidth]{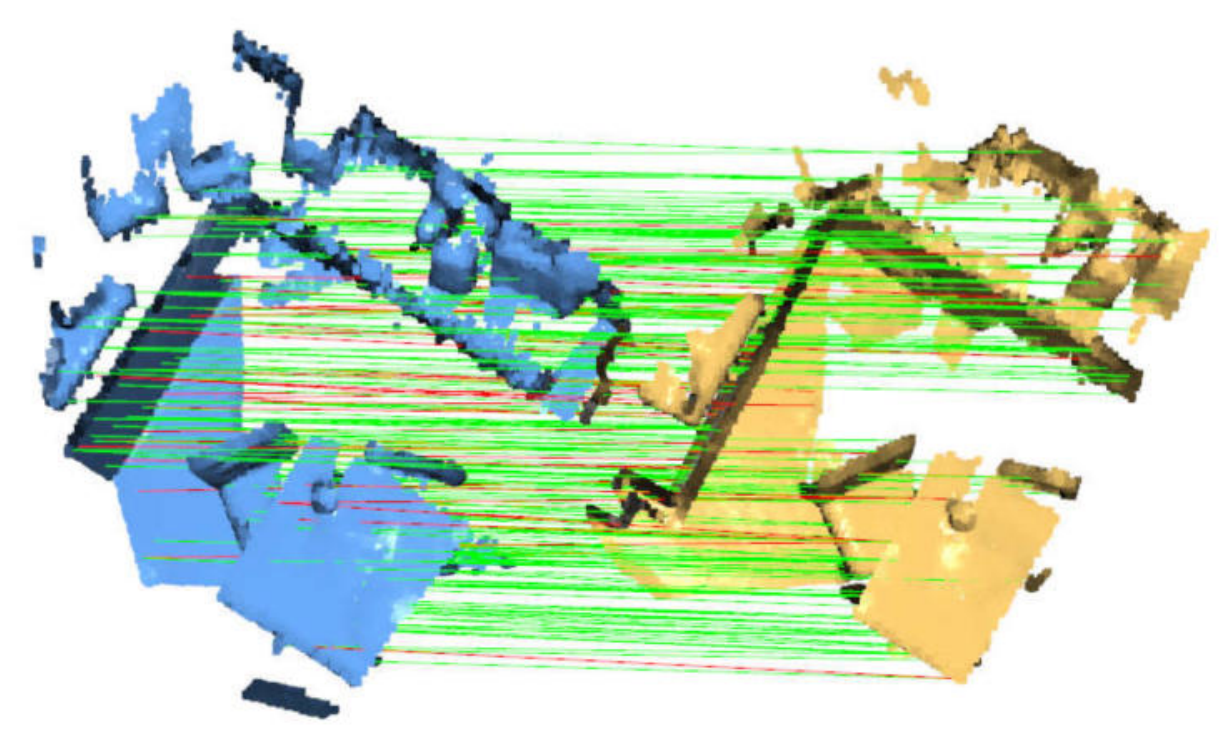}}
		\centerline{(d) hybrid point}
	\end{minipage}
	\caption{The visualization of patch correspondences on 3DMatch.  }
	\label{fig3}
\end{figure}

\textbf{Ablation results: }
Firstly to prove that hybrid points and dual-classes patch matching module work, we compare 1) adopting superpoints as nodes, 2) adopting non-salient points as nodes, 3) adopting salient points as nodes, 4) adopting hybrid points as nodes, and 5) adopting hybrid points as nodes while employing dual-classes patch matching module. As shown in Table \ref{tab2}, combining hybrid points and dual-classes patch matching could maximize the model's performance. The qualitative results can be seen in Fig.  \ref{fig3}.

Secondly, to justify only the non-salient branch need to use spectral matching(SM)~\cite{leordeanu2005spectral} in dual-classes patch matching, we compare the following manners of using SM block: 1) Applying SM block in both salient and non-salient branches. 2) Applying SM block in either salient branch or non-salient branch. 3) Applying SM block in neither the salient nor non-salient branches. As shown in Table \ref{tabb3}, using SM only in the non-salient branch provides the most significant performance boost for the model. The following analysis is introduced to explain why SM should be only used in non-salient branch.

The spectral method is widely used for retrieving the inliers from initial correspondences. SM can highlight the most reliable correspondences when the fraction of inliers is low and discard the others. However, if the fraction of inliers is high, SM may suppress many correspondences, which may not be as precise as the remaining ones but still can improve the estimated result. Moreover, if the feature matching is relatively accurate, forcibly using SM to highlight the most reliable ones will often cause the retained feature matches to concentrate in a small area and cannot consider the global optimal.

\subsection{KITTI odometry}
\textbf{Dataset: }KITTI odometry~\cite{geiger2012we} is an outdoor dataset consisting of 11 sequences of outdoor driving scenarios scanned by LiDAR. We follow ~\cite{bai2020d3feat, choy2019fully, huang2021predator} to split the dataset and refine the ground-truth transformation by ICP~\cite{schenker1992sensor}. Besides, only point cloud pairs that are at least 10m away are selected for evaluation.

\begin{table}[t]
\begin{center}
\setlength\tabcolsep{6pt}
\caption{ Registration results on KITTI odometry.}\label{tab4}
\begin{tabular}{c|cccc}
  \hline
  Methods &RR(\%)&RRE(\textdegree)&RTE(cm)\\
  \hline
  FCGF~\cite{choy2019fully}  &96.6&0.30&9.5\\
  D3Feat~\cite{bai2020d3feat} & 99.8 & 0.30 & 7.2\\
   Predator~\cite{huang2021predator} & 99.8 & 0.27 & 6.8\\
  CoFiNet~\cite{yu2021cofinet} & 99.8 & 0.41 & 8.2\\
  GeoTransfomer+LGR~\cite{qin2022geometric} &\underline{99.8}&\underline{0.24} & \underline{6.8} \\
  Ours&\bf{99.8}&\bf{0.22}&\bf{5.1}\\
  \hline
\end{tabular}
\end{center}
\end{table}
\textbf{Metrics: }
Following previous methods~\cite{huang2021predator, bai2020d3feat, choy2019fully}, our model is evaluated with three metrics: the Registration Recall (RR), the Relative Rotation Error (RRE), and the Relative Translation Error (RTE). Following \cite{bai2020d3feat, qin2022geometric}, the estimated matrix is considered accurate if the RRE is below 5 and RTE is below 2m.

\textbf{Registration results: }As shown in Table~\ref{tab4}, our model is compared with~\cite{choy2019fully, bai2020d3feat, huang2021predator, yu2021cofinet, qin2022geometric}.
For all the method, we use 1000 correspondences for evaluation. The result shows that our model achieves state-of-the-art performance in KITTI odometry. Compared to the previous method, our model has a lower RRE and RTE with a RR of 99.8\%, indicating that our model has better prediction precision.

\section{Conclusion}

In this paper, we propose to extract hybrid points as nodes for patch partition to highlight the prominent structure while considering
the completeness of point cloud description compared to the previous superpoints.
Furthermore, a dual-classes patch matching module is proposed for the hybrid points. 
We evaluate our method in 3Dmatch, 3DLomatch, and KITTI odometry datasets and conduct ablation experiments, showing that our method achieves state-of-the-art effects. 
In the future, we will combine semantic information to obtain a more consistent patch partition.

\section*{Acknowledgment}
This work was supported by the National Key Research and Development Program of China under Grant No. 2020AAA0108100.
\bibliographystyle{IEEEtran}


\end{document}